\documentclass{article}

\usepackage{PRIMEarxiv}

\usepackage[utf8]{inputenc} 
\usepackage[T1]{fontenc}    
\usepackage{hyperref}       
\usepackage{url}            
\usepackage{booktabs}       
\usepackage{amsfonts}       
\usepackage{amsmath}
\usepackage{amssymb}
\usepackage{nicefrac}       
\usepackage{microtype}      
\usepackage{lipsum}
\usepackage{fancyhdr}       
\usepackage{graphicx}       
\graphicspath{{media/}}     

\newcommand{\ts}{\textsuperscript}

\title{Data Generation for Satellite Image Classification Using Self-Supervised Representation Learning
}

\author{
  Sarun Gulyanon, Wasit Limprasert, Rachada Kongkachandra \\
  College of Interdisciplinary Studies \\
  Thammasat University \\
  Pathum Thani, Thailand\\
  \texttt{\{sarung, wasit, krachada\}email@tu.ac.th} \\
   \And
  Pokpong Songmuang \\
  Faculty of Science and Technology \\
  Thammasat University \\
  Pathum Thani, Thailand\\
  \texttt{pokpongs@tu.ac.th} \\
}

\begin{document}
\maketitle

\begin{abstract}
Supervised deep neural networks are the-state-of-the-art for many tasks in the remote sensing domain, against the fact that such techniques require the dataset consisting of pairs of input and label, which are rare and expensive to collect in term of both manpower and resources. On the other hand, there are abundance of raw satellite images available both for commercial and academic purposes. Hence, in this work, we tackle the insufficient labeled data problem in satellite image classification task by introducing the process based on the self-supervised learning technique to create the synthetic labels for satellite image patches. These synthetic labels can be used as the training dataset for the existing supervised learning techniques. In our experiments, we show that the models trained on the synthetic labels give similar performance to the models trained on the real labels. And in the process of creating the synthetic labels, we also obtain the visual representation vectors that are versatile and knowledge transferable.
\end{abstract}


\section{Introduction}
In the big data era, raw data are abundant, especially in remote sensing domain, where petabytes of satellite images (e.g., Sentinel-2 image data\footnote{https://www.sentinel-hub.com/}) are publicly available. In the meanwhile, the introduction of deep learning just over a decade ago gave us a tool to exploit the plentiful amount of data. Supervised learning becomes the state-of-the-art in many tasks \cite{Zhong2017a,Li2018,Gu2019} and been prevailed in many fields such as computer vision \cite{Russakovsky2015} and remote sensing \cite{Demir2018}. Supervised learning can solve any problem that can be described by pairs of the input and the corresponding expected 
 output, given that there are enough number of data pairs and enough computational power. One caveat is that the amount of the data and the computational power required to give good performance depends on the complexity of the task.

Hence, some recent researches focus on making the deep neural networks more efficient (e.g., \cite{Iandola2016,Tan2019}), usually by exploiting different neural network architecture with a lower number of parameters. This has two main benefits: faster computation and lower memory requirements, which enable the deployment of deep learning on mobile and embedded systems such as TensorRT, the deep learning inference optimizer, by NVIDIA \cite{NVIDIA2020}, TensorFlow.js for deploying deep learning in web application using javascript, and TensorFlow Lite \cite{Abadi2015} for deploying deep learning on IoT (internet of things) devices.

The issue of the dataset size is more subtle. The common tasks, e.g., generic object detection and localization, are not problematic because they can attract attentions of researchers and resources since they are well-known and they have plenty applications. So it is feasible to collect and annotate a large amount of samples for such tasks \cite{Russakovsky2015}. However, domain-specific or specialized tasks are more likely to suffer from this issue. The main challenge is the annotation process for a large number of samples and, for some tasks, the annotators may be required to have domain knowledge as well. When resources are limited, it is difficult to obtain the labels that are correct and consistent. Many previous works have been studied to tackle this issue. The full review of recent techniques will be discussed in Section~\ref{sec:litreview}.

In this work, we focus on the lack of annotation data problem for satellite image classification. In this task, a small number of labeled data and a large number of unlabeled data are given. We utilize the unlabeled data via self-supervised learning. Our process starts with finding the embedding vectors that capture the visual representations using the SimCLR \cite{Chen2020}, a simple contrastive learning framework. Next, a machine learning technique, i.e., support vector machine (SVM) \cite{Cortes1995}, is trained in the embedding space on the available labeled data. We use the result model to make the predictions for the unlabeled data, called the ``synthetic labels''\footnote{Some literatures call these predictions the pseudo labels, but the word has other meaning in the self-supervised context. So we opt for the synthetic labels here.}, which are used as the training dataset. Thus, our two main contributions are:

\begin{itemize}
	\item Introducing an automated process to produce the synthetic labels and demonstrating that a classification technique trained on the synthetic labels give similar results to a model trained on the real labels.
	\item Demonstrating that the visual representations learned from the unlabeled data can capture generic features and are versatile enough to be used in similar learning tasks with the same domain.
\end{itemize}

We tested our work on two satellite image datasets --- Amazon rainforest and oil palm plantation. Our experiments show that our method requires 9x less data to perform as well as a supervised learning. And the visual representations learned from the the larger dataset can be used to train the models for multiple learning tasks.

\section{Related work} \label{sec:litreview}
There are a large number of studies on how to incorporate the unlabeled data \cite{Schmarje2020} but we will focus on the techniques that are popular in remote sensing domain (i.e., data augmentation \cite{Abdelhack2020} and transfer learning \cite{Xie2016,GiorgianidoNascimento2020}) and the self-supervised learning, which is the foundation of our method, along with semi-supervised learning. Also the satellite image preprocessing plays an important role in obtaining good performance (e.g., haze removal \cite{He2011}, super-resolution \cite{Ghaffar2019,Shermeyer2019}) but it is task specific; for more details, refer to \cite{Schowengerdt2012,Sowmya2017}.

\subsection{Data Augmentation}

It is a common knowledge that the more data a machine learning algorithm, especially deep learning, has for training, the more likely it will perform better. The reason is that these networks have a huge number of parameters so they have a tendency to overfit. Exposing these networks to a variety of data helps avoid overfitting and increase generalization. Hence, the intuition of data augmentation is to generate more data using the existing samples with labels. Data Augmentation approaches tackle the training dataset directly. It operates under the assumption that more information can be extracted from the original data through augmentations \cite{Perez2017,Shorten2019}.

Previous work has demonstrated the effectiveness of data augmentation, especially in computer vision field via geometric transformation, such as cropping, rotating, translation, scaling, and flipping input images. One caveat is that the applied transformation must preserve the original label of the sample. One notable neural network architecture that exploits this technique is the ``AlexNet'' \cite{Krizhevsky2012}, resulting in the breakthrough in computer vision. Color transformation is also a popular practice since images are usually represented by color pixels so perturbing colors creates different images from computer perspective. Or random erasing \cite{Zhong2017} is another new technique to tackle the occlusion problem, that occurs when parts of the object are unobservable. Random erasing randomly selects a patch in an image and masks it with some constant or random values.

Another strategy for data augmentation in computer vision is through adversarial training, commonly through the generative adversarial networks (GANs) to generate images of different styles from the same types. GANs consist of two groups of networks --- generator and discriminator --- with contrasting objectives. The generator generated the data from some forms of input, which could be a random vector or another image; while, the discriminator distinguish the original samples from the generated samples. Thus, the generator from this framework has the ability to generate new training data \cite{Shorten2019}. However, the downfall of this framework is that it is difficult to train GANs in practice \cite{Mescheder2018}. High-resolution outputs produced by the generator is likely to cause training instability and non-convergence. And GANs require a substantial amount of data to train \cite{Salimans2016}. Hence, these properties make it infeasible to apply GANs on small dataset.

\subsection{Transfer Learning}
Transfer learning addresses the insufficient training data by transferring the knowledge from other tasks with the same or similar input domains that already exists. In order to be able to use this technique, we must relax a major assumption in many machine learning algorithms that the training data must be independent and identically distributed (i.i.d.) with the test data. Following this assumption, when the task is changed, the distribution changes and the machine learning models must be rebuilt from scratch using the new training data. Relaxing this assumption allows us to improve the predictive function of the learning task ${\mathcal T}_T$ in a target domain ${\mathcal D}_T$ using the given learning task ${\mathcal T}_S$, ${\mathcal T}_S \neq {\mathcal T}_T$, in a source domain ${\mathcal D}_S$, ${\mathcal D}_S \neq {\mathcal D}_T$ \cite{Pan2010}.

Transfer learning can utilize the knowledge from other domains and tasks at different levels. Transfer learning techniques can be grouped into two categories based on the transferred knowledge: sample-based and latent-feature-based. Sample-based techniques utilize instances directly; while, latent-feature-based techniques find the feature extraction function and utilize the function itself or the result feature vectors.  Different types of techniques make a trade-off between complexity of the technique, adaptability to new domains, and flexibility to new neural architectures.

Sample-based techniques have two subgroups: instances-based and mapping-based. Instances-based techniques assign the instances from the source domain as supplements to the training set in the target domain along with the appropriate weight values to the selected instances. This strategy can be applied only if the instances in source domain compose a distribution similar to the target domain, where the weight value depends on this similarity. Meanwhile, mapping-based techniques rely on the mapping to find the instances from the source domain that are similar to the instances from the target domain in a new data space \cite{Tan2018}.

Latent-feature-based techniques exploit the solution models, which usually are deep neural networks, for extracting the features, called transferable features, the features that can be applied in the learning task in the target domain. There are two subgroups: network-based and adversarial-based. Network-based techniques utilize the partial of the neural network that was trained for the learning task in the source domain beforehand to extract the transferable features. This type of techniques is based on the assumption that the higher layers of the neural network capture the higher level of abstraction. Hence, the lower layers capture the generic features that are versatile. These techniques divide the neural network into two parts, the feature transformer and classifier. The feature transformer is preserved, while the classifier is changed to match the learning task. Meanwhile, adversarial-based techniques exploit GANs to extract transferable features, unlike the data augmentation that exploits GANs to produce new training data. Good transferable features must be discriminative for the target learning task but indiscriminate between the input domains \cite{Pan2010,Weiss2016,Tan2018}.

\subsection{Semi-Supervised Learning}
Semi-Supervised Learning is a special form of classification. Traditional classification task requires the labeled data, where the data collection is an expensive and error prone process, especially the label annotation step. Meanwhile unlabeled data are often relatively cheap to collect. Semi-supervised learning exploits a large amount of unlabeled data, together with a much smaller number of labeled data to build better classifiers \cite{Zhu2005}.

There is a wide literature on semi-supervised learning techniques. The comprehensive overviews are in \cite{Zhu2005,Chapelle2006,VanEngelen2020}. Semi-supervised techniques can be categorized into two types based on their objectives: inductive and transductive methods. Inductive methods aim to find a predictor function that maps any data point in the input space to a label prediction; while, transductive methods focus on obtaining label predictions for the given unlabeled data points without involving the predictor function \cite{VanEngelen2020}. Here, we will focus on one particular type of inductive methods, the wrapper methods, which our method is based on.

Wrapper method is a simple semi-supervised learning technique that incorporate the unlabeled data in a straightforward manner. This approach extends an existing supervised algorithms to semi-supervised learning in two steps: training and pseudo-labeling. In the training step, we train a classifier, called the wrapper method, on labeled data. The pseudo-labeling step uses the wrapper method to create the synthetic labels for unlabeled data. These two steps can be iterative. Then, the classifiers can be re-trained on the labeled data as well as the unlabeled data with synthetic labels to create the final inductive classifier. Hence, the real labels and the synthetic labels are indistinguishable to the inductive classifier so this approach can be used with any supervised base learner \cite{VanEngelen2020}. 

\subsection{Self-Supervised Learning}
Self-supervised learning is a type of unsupervised learning methods but it formulates the problem as if it is a supervised learning. The key difference is that the labels are automatically generated, called the pseudo labels, and the labels can be irrelevant to the target learning task. The general pipeline of self-supervised learning consists of two main steps --- pretext task training and downstream task training \cite{Jing2019a}. 

Pretext tasks refer to the pre-designed tasks that help generate useful visual representation features.
The pretext task training is formulated as the supervised learning that aims to predict the pseudo labels, which are generated based on some attributes of both labeled and unlabeled data. The purpose of this step is not to find a predictor but to find a quality visual representation features. 

The downstream tasks refer to tasks that are used to evaluate the quality of visual representation features from the first step. The downstream task training transfers the learned features to the downstream tasks and train a supervised model in the visual representation space on the labeled data (even if the data is relatively small). The quality of visual representation features is reflected by the performance of the model on downstream task.

One notable type of techniques in self-supervised is contrastive visual representation learning. The idea was introduced by Hadsell et al. in 2006 \cite{Hadsell2006}. It addresses the representation learning by the pretext task of contrasting positive pairs of samples against negative ones. The authors in \cite{Dosovitskiy2014} propose the pretext task of discriminating each instance, which is augmented to create a surrogate class. The method in \cite{Chen2020,Chen2020a}, called SimCLR, presents a simple framework for contrastive learning of visual representations based on the pretext task of maximizing agreement between differently augmented views of the same sample via a contrastive loss in the non-linear latent space.

\begin{figure*}[t]
	\centering
	\includegraphics[width=0.595\textwidth]{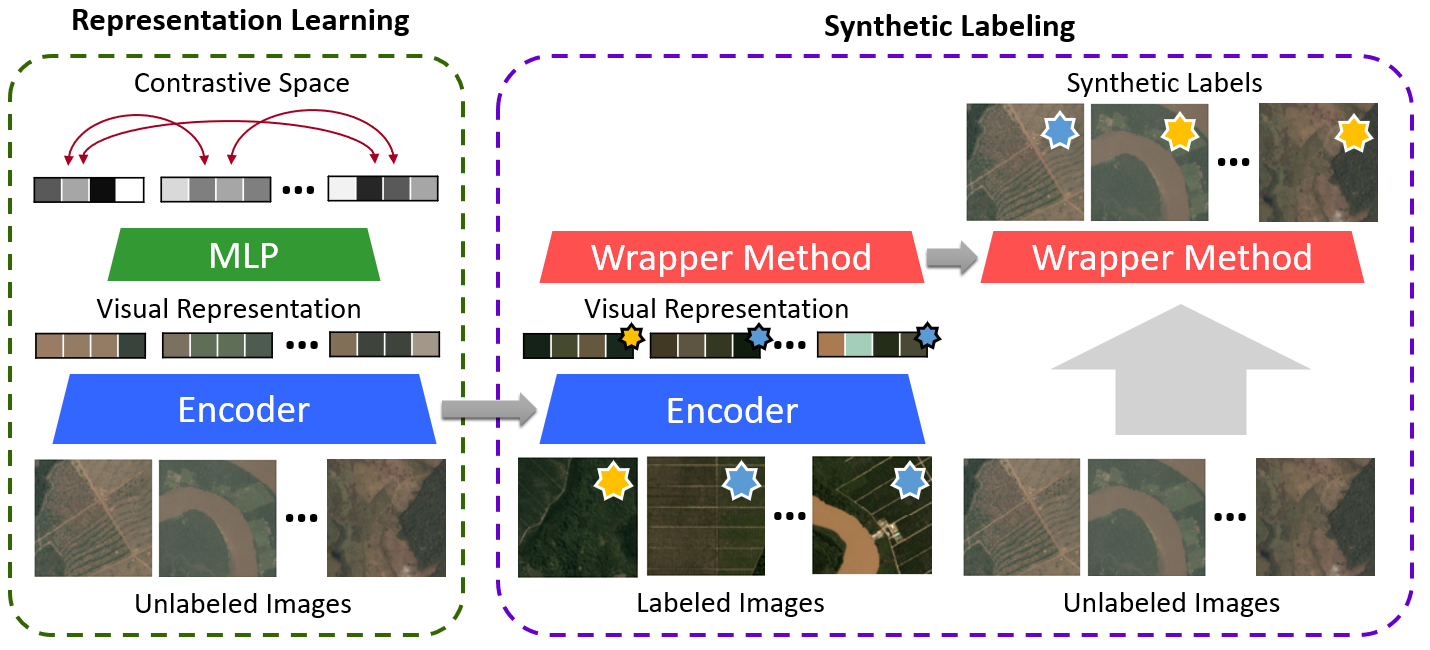}
	\includegraphics[width=0.39\textwidth]{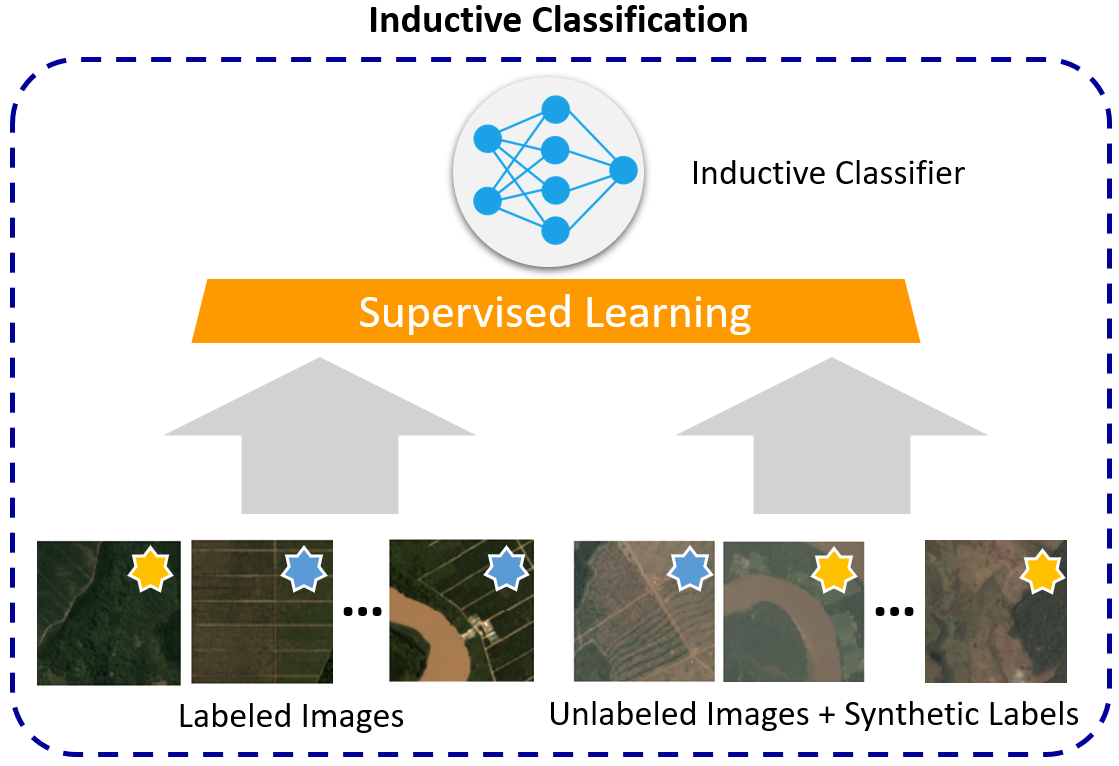}
	\caption{shows the overview of our method that comprises a sequence of three steps showed in colored dashed boxes: representation learning, synthetic-labeling, and inductive classification. (Left) the representation learning step learns the visual representation encoder. (Middle) the synthetic labeling utilize the previously learned encoder to generate the synthetic labels for unlabeled data. (Right) the inductive classification trains a supervised model on both the training set and the unlabeled set with their corresponding synthetic labels.}
	\label{fig:method}
\end{figure*}

\section{Method}
The contrastive visual representation learning introduces a new kind of supervised base learner for semi-supervised wrapper methods, from which it removes the necessity of iteration. Our method approaches the synthetic label data generation task for satellite images using the wrapper method with SimCLR \cite{Chen2020} as the base learner. 

Let ${\mathcal{D}} = \{ \text{x}_i \}$, for $i = 1,...,N$, denote a set of unlabeled satellite images and $D = \{ (x_j, y_j) \}$, for $j = 1,...,M$, denote a set of labeled satellite, where $N \gg M$. Given the unlabeled data ${\mathcal{D}}$ and labeled data $D$, the synthetic label data generation task is to find the synthetic labels, $\text{y}_i$, of the input $\text{x}_i$.

Our method consists of three steps: representation learning, synthetic-labeling, and inductive classification (Fig.~\ref{fig:method}).

\subsection{Representation Learning}
The aim of the first step is to find the embedding vectors that capture the quality visual representations. We adopted the self-supervised learning method, SimCLR \cite{Chen2020}, to generate the embedding vectors that represent the visual meaning from both unlabeled ${\mathcal{D}}$ and labeled data $D$.

The embedding vectors ${\bf h}_i$ are generated by applying the encoder network $f$ on the augmented data:
\begin{equation}
{\bf h}_i = f(t(x_i), \theta) 
\end{equation}
where $\theta$ is the parameter of $f$ and $t$ is the stochastic data augmentation transformation.

The embedding vectors are projected to the vector ${\bf z}_i$ in the contrastive space for contrastive learning.
\begin{equation}
{\bf z}_i = g({\bf h}_i, W)
\end{equation}
where $g$ is the multi-layered perceptron (MLP) with one hidden layer and non-linear activation function. $W$ are the parameters of $g$.

The objective function is to minimize the contrastive loss function, called NT-Xent loss \cite{Chen2020}, which maximizing agreement between differently augmented views of the same data,
\begin{equation}
\theta^* = \arg \min_{\theta, W} {\mathcal{L}}
\end{equation}
\begin{equation}
{\mathcal{L}} = \frac{1}{2N} \sum^N_{k=1}  \left[ l ( 2 k-1,2k) + l(2k, 2k-1) \right]
\end{equation}
\begin{equation}
l(i,j) = -\log \frac{\exp(sim ({\bf z}_i, {\bf z}_j) / \tau)} { \sum^{2N}_{k=1} {\bf {1}}_{[k \neq i]} \exp(sim ({\bf z}_i, {\bf z}_j) / \tau) }
\end{equation}
where $sim({\bf z}_i, {\bf z}_j)$ denotes the cosine similarity between two vectors ${\bf z}_i$ and ${\bf z}_j$.
${\bf {1}}_{[k \neq i]} \in \{0, 1\}$ is an indicator function evaluating to 1 iff $k \neq i$. The loss is computed across all positive pairs only, both $({\bf z}_i, {\bf z}_j)$ and $({\bf z}_j, {\bf z}_i)$, in a batch \cite{Chen2020}. 

\begin{figure*}[t]
	\centering
	\includegraphics[width=0.47\textwidth]{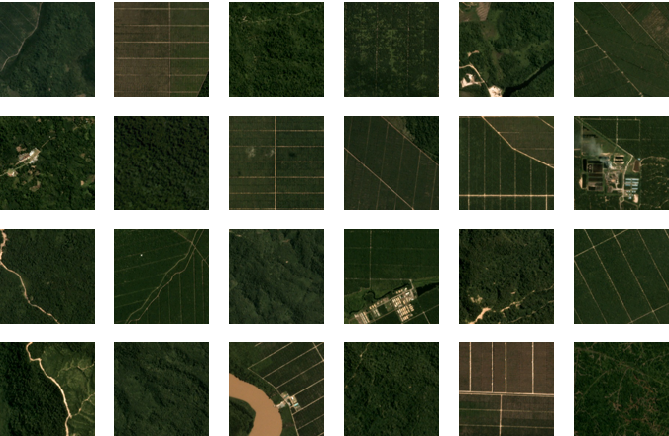}
	\quad
	\includegraphics[width=0.47\textwidth]{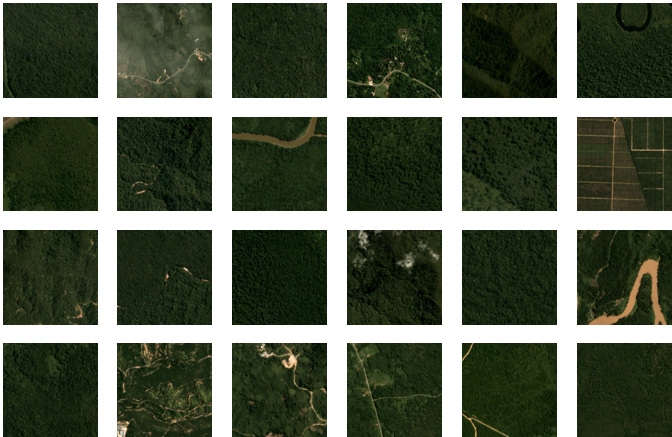}
	\caption{shows sample images from palm oil plantation dataset. (Left) the images that contain palm oil plantation(s). (Right) the images that don't contain palm oil plantation.}\label{fig:oilpalmdataset}
\end{figure*}

\begin{figure*}[t]
	\centering
	\includegraphics[width=0.47\textwidth]{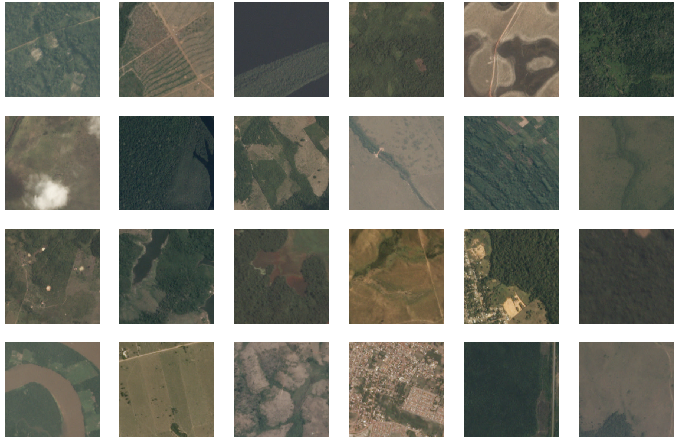}
	\quad
	\includegraphics[width=0.47\textwidth]{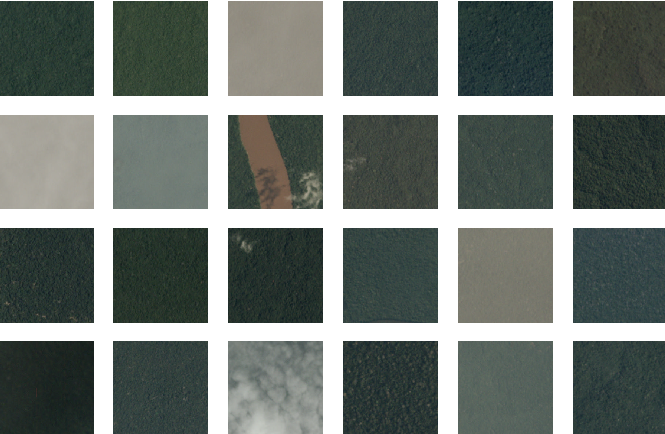}
	\caption{shows sample images from Amazon rainforest dataset. (Left) the images with agriculture label. (Right) the images without agriculture label.}\label{fig:amazondataset}
\end{figure*}

\subsection{Synthetic Labeling}
The goal of this step is to generate the synthetic labels of the unlabeled data ${\mathcal D}$. First, a wrapper method $\mathcal{F}$ is trained in the embedding space $\{ ({\bf h}_j, y_j) \}$ over the available labeled data $D$. Then, we use the wrapper model to generate the synthetic labels of unlabeled data,
\begin{equation}
\text{y}_i = { \mathcal{F} }(\text{x}_i)
\end{equation}

The wrapper method should be non-linear model since it was empirically shown that non-linear projection gives better performance \cite{Chen2020}. Similarly, non-linear models also improve the knowledge transfers to the downstream task.

\subsection{Inductive Classification}
This last step is intent to construct the final inductive classifier using both labeled data, $D$, and unlabeled data with synthetic labels, $\{ \text{x}_i, \text{y}_i \}$, as the training data. Using self-supervised learning as the wrapper method eliminates the needs for repeatedly perform the pseudo-labeling on unlabeled data, which helps avoid expensive computations and convergence problems.

A choice on the inductive classifier is arbitrary. It does not need to be the same as the wrapper method. This is a useful property, as described in \cite{Hinton2015}, since the model appropriate for deployment can be used regardless of the visual representation learning.

\section{Experiments}

We validated our method using two datasets --- Amazon rainforest and palm oil plantation datasets. Both datasets contain RGB image patches of satellite images of size 256$\times$256 pixels. Although, the data in these datasets are all labeled, we simulate the scenarios where labeled data are scarce like tasks in \cite{Kussul2017,Heupel2018} by sampling a part of the datasets and dropping the labels form the unlabeled set. The sampling is done instead of collecting new unlabeled data to avoid data issues and preprocessing complications.

\begin{itemize}
	\item {\bf Palm oil plantation dataset} are 3m resolution RGB images gathered by the Planet's satellites. The label indicates the presence or absence of oil palm plantation in the image chip. The labels come from the crowdsourced annotations. The dataset contains over 15,000 patches. We sample only 1,885 images as the labeled data, with 918 as oil palm plantation tag\footnote{WiDS Datathon 2019 \\ https://www.kaggle.com/c/widsdatathon2019} (Fig.~\ref{fig:oilpalmdataset}).
	
	\item {\bf Amazon rainforest dataset} are 4-bands satellite images at 3m resolution of the Amazon basin. The data were collected by Planet's Flock 2 satellites between January 1\ts{st}, 2016 and February 1\ts{st}, 2017. The dataset contains over 40,000 patches with multilabels indicating atmospheric conditions and land use phenomena. In this work, we sample only 21,118 images as the labeled data from JPG chips of RGB images, instead of the raw 4-bands TIF images, with only one tag, the agriculture label, to obtain the binary classification problem. The sampling data contain 9,852 agriculture tags\footnote{Planet: Understanding the Amazon from Space https://www.kaggle.com/c/planet-understanding-the-amazon-from-space} (Fig.~\ref{fig:amazondataset}).	
\end{itemize}

All these experiments were executed on a computer with AMD Ryzen 7 2700X 3.7GHz processor, GeForce Nvidia RTX 2070, and RAM 16 GB. Training all three steps take a few hours.

Our experiments aim to validate our method mainly in two fronts: the performance of the inductive classifier, and the quality and versatility of the visual representation vectors.

\begin{figure*}[t]
	\centering
	\includegraphics[width=0.4\textwidth]{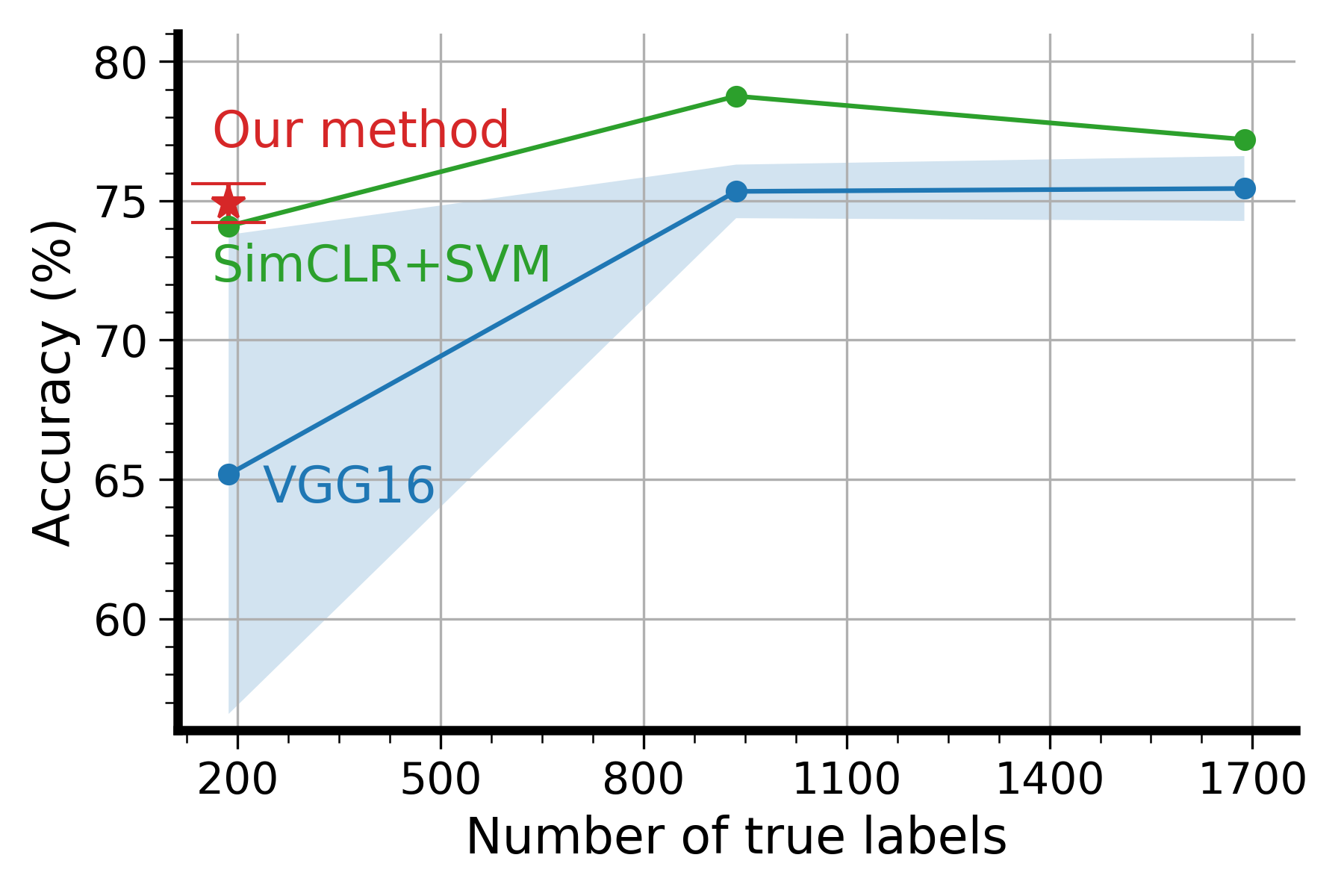}
	\qquad
	\includegraphics[width=0.4\textwidth]{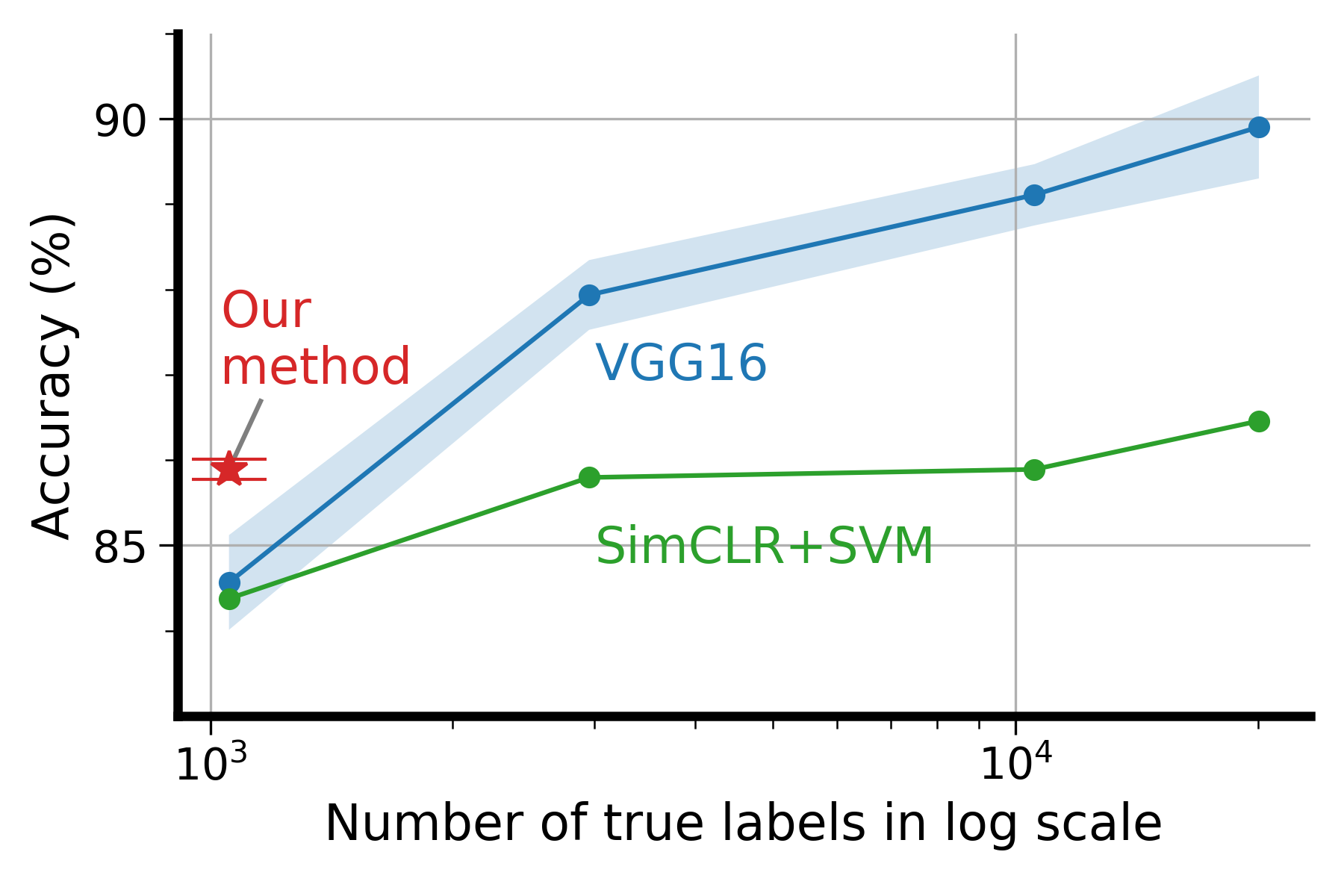}
	\caption{compares the accuracy of different methods in colors: our method (red), pre-trained VGG16 model (blue), and SVM over SimCLR features (green). The error bar and highlighted area show the range of one standard deviation of the mean from running the algorithm 5 times. The X-axis shows the number labeled data available to the algorithm. (Left)  the result from Palm oil plantation dataset (Right) the result from Amazon rainforest dataset.}
	\label{fig:results}
\end{figure*}

\subsection{Inductive Classifier Performance}
First, we show that the synthetic labels can be used as a substitution in the absence of real labels. We validate our method by evaluating the pre-trained convolutional neural network (CNN), i.e., VGG16 \cite{Simonyan2014}, trained on the synthetic labels. VGG16 is a popular choice, which works well with the small datasets that we used in the experiements. More sophisticated models can be used here but they usually require larger datasets and more computational power. As a showcase of our method, we opt for a simple model. The synthetic labels are generated using the SVM over SimCLR features trained on labeled data.

\begin{table*}[tb!]
	\begin{center}
		\begin{tabular}{ |l|c|c|c|c| } 
			\hline
			&{\small Accuracy} & {\small Precision} & {\small Recall} & {\small F1-score} \\ \hline
			Ours 	& \bf{ 74.9} & \bf{ .798} & .643 & \bf{ .712}  \\
			VGG16	& 65.2 & .708 & .647 & .636  \\ 
			SimCLR 	& 74.1 & .772 & \bf{ .656} &  .709 \\ 
			\hline
		\end{tabular}
	\end{center}
	\caption{compares the average accuracy between the methods when 187 labeled data from palm oil plantation dataset are available.}
	\label{table:result_palmoil}
\end{table*}

\begin{table*}[tb!]
	\begin{center}
		\begin{tabular}{ |l|c|c|c|c| } 
			\hline
			&{\small Accuracy} & {\small Precision} & {\small Recall} & {\small F1-score} \\ \hline
			Ours 	& \bf{ 85.9} & \bf{ .897} & .831 & \bf{ .863}  \\
			VGG16	& 84.6 & .851 & \bf{ .862} & .856  \\ 
			SimCLR 	& 84.4 & .819 & .854 & .836 \\ 
			\hline
		\end{tabular}
	\end{center}
	\caption{compares the average accuracy between the methods when 1,055 labeled data from Amazon rainforest dataset are available.}
	\label{table:result_amazon}
\end{table*}

We compare our method against 1) pre-trained VGG16 model (VGG16) and 2) SVM over SimCLR features (SimCLR). Transfer learning is currently the state-of-the-art for some remote sensing tasks \cite{GiorgianidoNascimento2020}. The VGG16 model is initialized by the pre-trained ImageNet weights \cite{Russakovsky2015} and the input data are augmented by shearing and flipping both horizontally and vertically. The SVM with radial basis function (RBF) kernel for non-linearity is trained over the representation vectors learned by SimCLR. SimCLR is trained with ResNet18 \cite{He2016} as the encoder network.

The only image preprocessing step that is done in all experiments is scaling to 128$\times$128 pixels for memory efficiency. Although, image preprocessing has a significant impact on the performance \cite{Sowmya2017}, it is not the focus of this work so we opt for the simplest one. To avoid imbalanced data issue, we sample and split the dataset so that both training and test sets have similar number of labels. In the palm oil dataset has 193 images in the test set, with 93 palm oil plantation images. In the Amazon rainforest dataset contains 1,056 images, with 493 agriculture tags.

\begin{table}[t!]
	\begin{center}
		\begin{tabular}{ |l|c|c| } 
			\hline
			&{\small Accuracy} & {\small CNN Accuracy}  \\ \hline			
			SVM		& 74.1 & 74.9 $\pm$ 0.7 \\ 
			kNN 	& 70.5 & 76.2 $\pm$ 0.9 \\ 
			Logistic& 57.5 & 71.2 $\pm$ 2.0 \\
			\hline
		\end{tabular}
	\end{center}
	\caption{evaluates the synthetic labels quality by comparing different wrapper methods and the CNN models trained on their synthetic labels. 187 labeled data from palm oil plantation dataset are available.}
	\label{table:synthetic_label_palmoil}
\end{table}

\begin{table}[t!]
	\begin{center}
		\begin{tabular}{ |l|c|c| } 
			\hline
			&{\small Accuracy} & {\small CNN Accuracy}  \\ \hline
			SVM		& 84.4 & 85.2 $\pm$ 0.2 \\ 
			kNN 	& 83.0 & 84.9 $\pm$ 0.5 \\ 
			Logistic& 79.2 & 83.7 $\pm$ 0.5 \\
			\hline
		\end{tabular}
	\end{center}
	\caption{evaluates the synthetic labels quality by comparing different wrapper methods and the CNN models trained on their synthetic labels. 1,055 labeled data from Amazon rainforest dataset are available.}
	\label{table:synthetic_label_amazon}
\end{table}

\begin{table*}[t!]
	\begin{center}
		\begin{tabular}{ |l|c|c|c|c| } 
			\hline
			&{\small Accuracy} & {\small Precision} & {\small Recall} & {\small F1-score} \\ \hline
			VGG16	& 65.2 $\pm$ 8.6 & .708 $\pm$ .132 & .647 $\pm$ .203 & .636 $\pm$ .049 \\ 
			Transferred SVM	& 72.5 & .692 & .774 & .731  \\ 
			Transferred CNN & 73.2 $\pm$ 0.7 & .716 $\pm$ .009 & .733 $\pm$ .004 & .725 $\pm$ .006 \\ 
			\hline
		\end{tabular}
	\end{center}
	\caption{compares the methods when 187 labeled data from palm oil plantation dataset are available.}
	\label{table:transfer}
\end{table*}

Tables~\ref{table:result_palmoil} and~\ref{table:result_amazon} shows the superiority of our method compared to the other methods when the number of labeled data is small. Our method can give nearly 10\% increase in accuracy compared to the pre-trained model, which is the state-of-the-art for many remote sensing tasks. Fig.~\ref{fig:results} (left) also shows that our method gives similar performance to the pre-trained model when the labels of all data are accessible, which means our method needs 9x less number of labels to perform as well as the pre-trained VGG16 model. Since CNN algorithm contains stochastic steps so we report the average result from 5 runs. This advantage diminishes when the number of labeled data is high as shown in Fig.~\ref{fig:results} (right), but in specialized domains, collecting thousands of labeled data is a very challenging and if a large number of labeled data can be obtained then the data is not the issue.

Second, we measure the effect of synthetic label quality by training CNN, i.e., pre-trained VGG16, on the synthetic labels generated by different wrapper techniques, i.e., 1) SVM, 2) $k$-nearest neighbor (kNN), and 3) logistic regression. We also measure the performance of these wrapper techniques on the test dataset as well. The results in Tables~\ref{table:synthetic_label_palmoil} and~\ref{table:synthetic_label_amazon} show that the CNN trained on the synthetic labels give higher accuracy. Moreover, the non-linear classifiers (i.e., SVM with RBF kernels and kNN) tends to work better than the linear classifier (i.e., logistic regression).

\begin{figure*}[t!]
	\centering
	\includegraphics[width=0.47\textwidth]{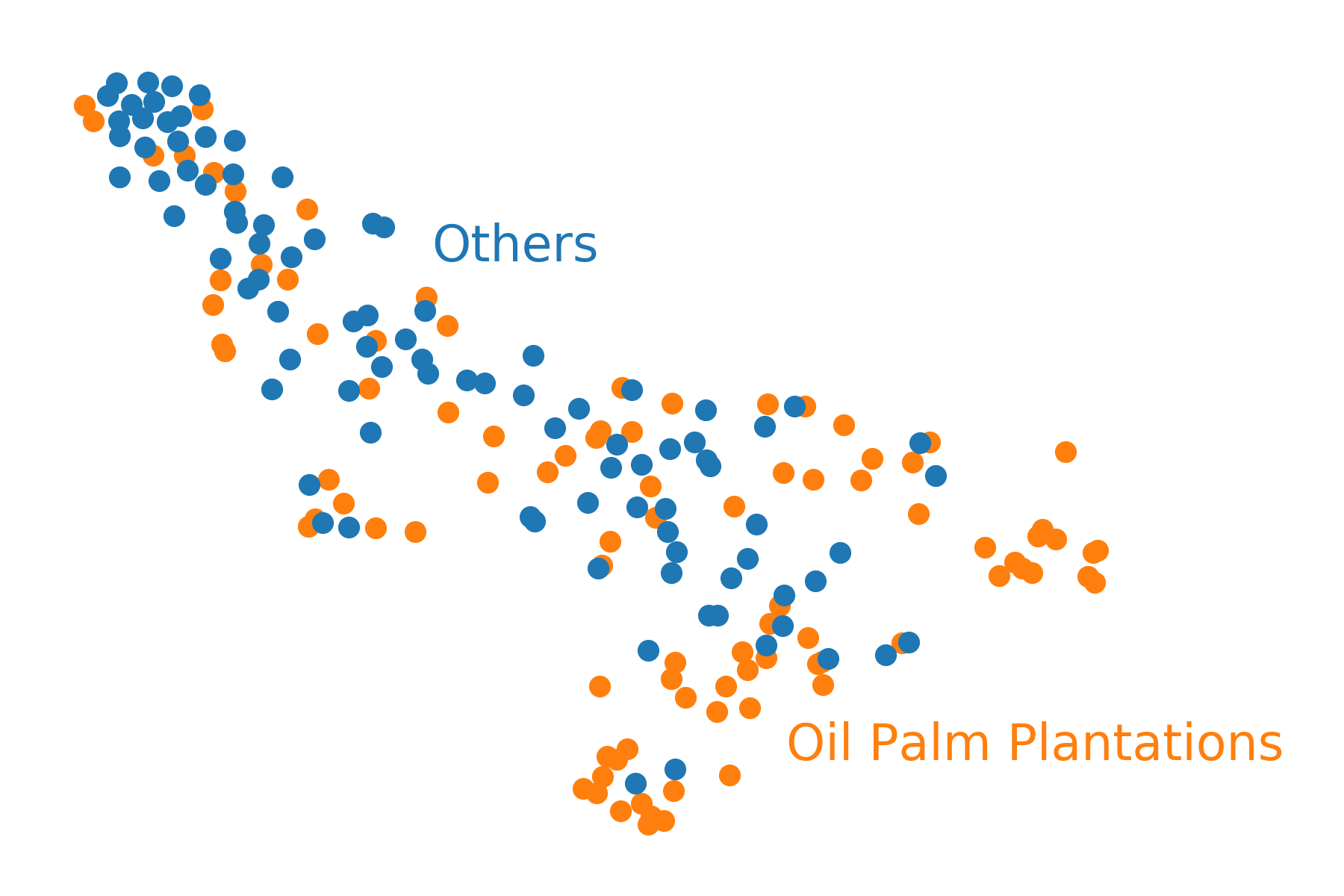}
	\qquad
	\includegraphics[width=0.47\textwidth]{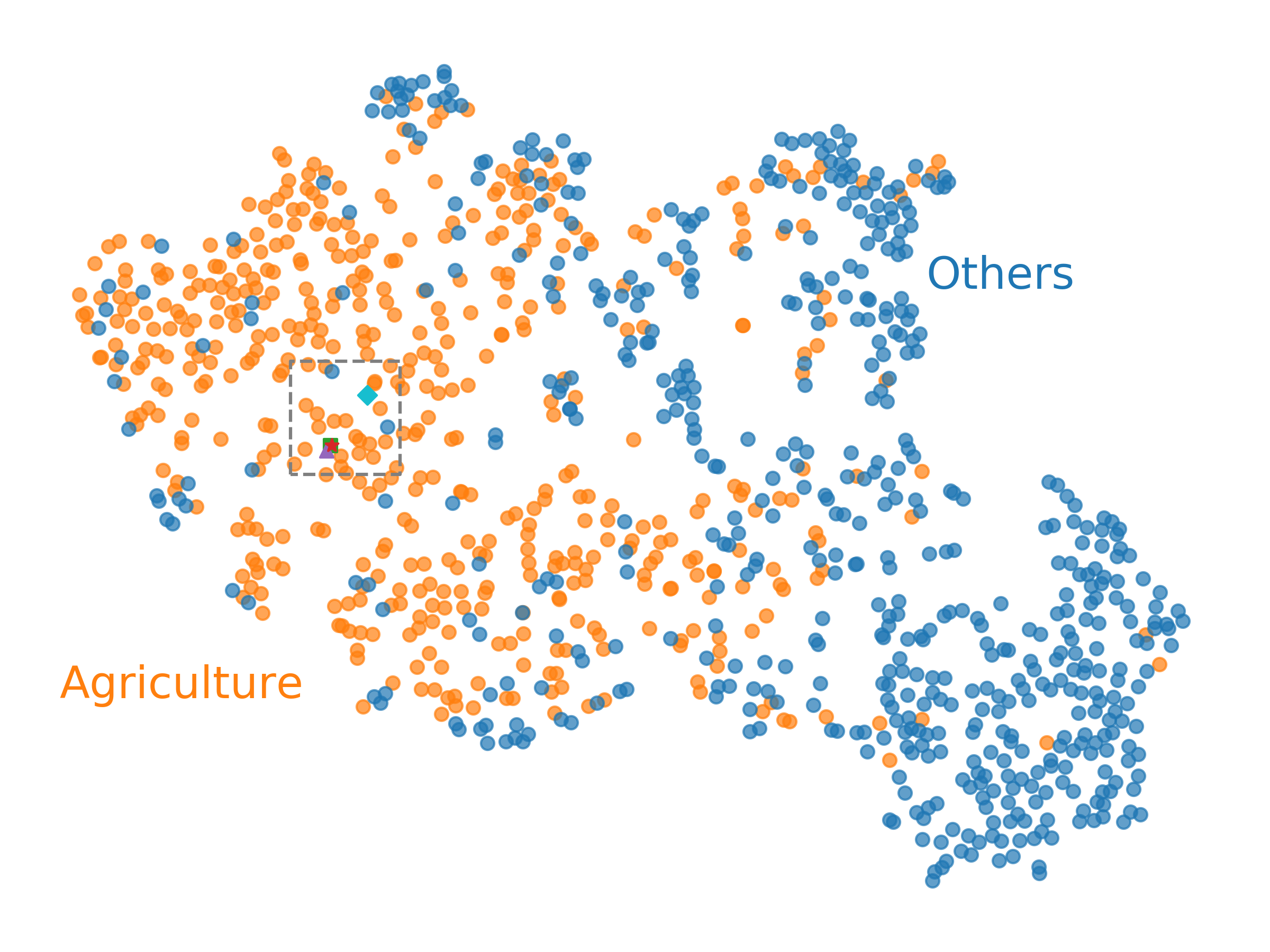}
	\caption{the t-SNE visualization of visual representation vectors of training set. The encoder function is optimized via self-supervised learning. (Left) Oil palm plantation training set. (Right) Amazon rainforest training set.}
	\label{fig:tsne}
\end{figure*}
\begin{figure}[t!]
    \centering
	\includegraphics[width=0.3\linewidth]{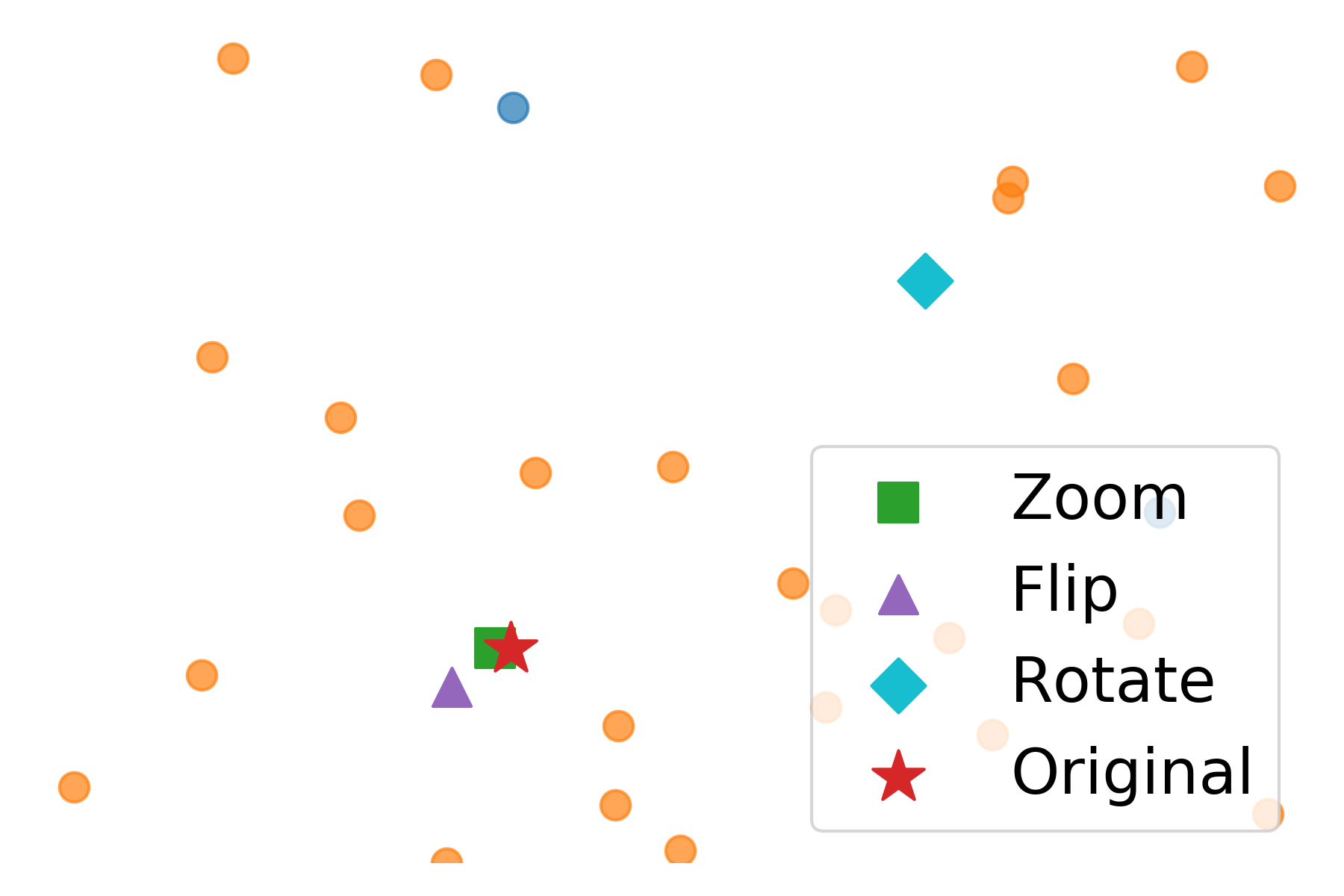}
	\\
	\includegraphics[width=0.15\linewidth]{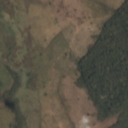}
	\includegraphics[width=0.15\linewidth]{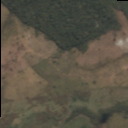}
	\includegraphics[width=0.15\linewidth]{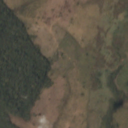}
	\includegraphics[width=0.15\linewidth]{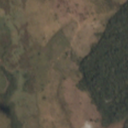}	
	\caption{show that the visual representation vectors are quite invariant under data augmentation. (Top) the area inside the gray dashed box in Fig~\ref{fig:tsne} (right). A sample is augmented by flipping, 90$^{\circ}$ rotation, and cropping. (Bottom) from left to right, original image, 90$^{\circ}$ rotated image, flipped image, and cropped image.}
	\label{fig:augment_tsne}
\end{figure}

\subsection{Representation Vector Quality}
Showing the quality of representation vector is more subtle. We saw the potential of these representation vectors in the previous sections, but this section will explore the capability of these representation vectors further. First, we show that the visual representation features learned through self-supervised learning work well as they show noticeable differences between classes through the t-SNE visualization \cite{VanDerMaaten2008} (Fig~\ref{fig:tsne}). In the palm oil plantation dataset, the difference is small as seen in Fig.~\ref{fig:oilpalmdataset}. On the other hand, the images of agriculture areas in Amazon rainforest dataset in Fig.~\ref{fig:amazondataset} are more apparent than the previous dataset; therefore, the separation between the two classes is more prominent in the visualization.

Another important observation is that the visual representation vectors capture some semantic features as they are invariant to some kinds of data augmentation. In Fig.~\ref{fig:augment_tsne}, we tested against 3 types of transformations --- rotation, zooming/cropping, and mirroring. We found that the embedding vectors of the transformed images are fairly similar to the embedding vector of the original image. This is a useful property since the representation will treat the transformed images the same way as the original image.

Second, we will show that the visual representation space is sufficiently generic for knowledge transfers. We use the encoder network learned from the Amazon rainforest dataset to embed the data in the oil palm plantation dataset. Here we trained the SVM as the wrapper method over the training set of palm oil plantation dataset embedded using the visual representation space learned from the Amazon rainforest dataset. Then, the transferred synthetic labels are generated by using the SVM to make the predictions over unlabeled data. Following our method, we train the pre-train VGG16 model over the transferred synthetic labels as well as the training set.

The result in Table~\ref{table:transfer} shows that both transferred wrapper method (Transferred SVM) and transferred inductive classifier (Transferred CNN) outperform the conventional pre-trained model (VGG16). This shows that the representation vectors are versatile.

These experiments show that the visual representations learned via self-supervised learning has many desirable properties from encoding the semantic features and the ability to enable knowledge transfers. Hence, our method combines this technique with the semi-supervised learning framework to give an even more superior result.

\section{Conclusion}
We presented the synthetic label data generation method that improves any supervised learning methods in the presence of insufficient labeled data issue as if they were trained on the labeled data. Our method combines the self-supervised representation learning with the semi-supervised framework to rely less on labeled data. Visual representation learning helps us learn the semantic features without the needs for any label. Meanwhile, semi-supervised learning helps oriented the visual representation space towards the target learning task. Hence, the self-supervised visual representation is knowledge transferable. We evaluated our method in satellite image classification task on two datasets of different size and objective, and our method shows superior results against the conventional solutions, i.e., transfer learning solutions.







\section*{Acknowledgments}
This work is supported by the Thailand Research Fund grant number RDG61Q0020.

\bibliographystyle{unsrt}

\end{document}